\documentclass[a4paper,fleqn]{cas-sc} 

\usepackage[authoryear,longnamesfirst]{natbib}
\usepackage{times}
 \usepackage[english]{babel}
\usepackage{setspace}
\usepackage{caption}
\usepackage{url}
\usepackage{color}
\newcommand{\ie}{\mbox{\textit{i.e.}, }}
\newcommand{\eg}{\mbox{\textit{e.g.}, }}
\usepackage{float}
\usepackage{amsmath,amssymb,amsfonts}

\usepackage{booktabs}
\usepackage{siunitx}

\usepackage{algpseudocode}
\usepackage[linesnumbered,ruled]{algorithm2e}
\usepackage{tabularx}
\usepackage{natbib}
\newtheorem{definition}{Definition}[section]

\makeatletter
\def\set@curr@file#1{%
  \begingroup
    \escapechar\m@ne
    \xdef\@curr@file{\expandafter\string\csname #1\endcsname}%
  \endgroup
}
\def\quote@name#1{"\quote@@name#1\@gobble""}
\def\quote@@name#1"{#1\quote@@name}
\def\unquote@name#1{\quote@@name#1\@gobble"}
\makeatother
\usepackage{graphicx}
\newcommand{\xmeno}{x^-}

\def\tsc#1{\csdef{#1}{\textsc{\lowercase{#1}}\xspace}}
\tsc{WGM}
\tsc{QE}
\tsc{EP}
\tsc{PMS}
\tsc{BEC}
\tsc{DE}

\begin{document}
\let\WriteBookmarks\relax
\def\floatpagepagefraction{1}
\def\textpagefraction{.001}
\shorttitle{Spillover Algorithm}
\shortauthors{M. Lujak, A. Fernandez, and E. Onaindia}

\title[mode = title]{Spillover Algorithm: A Decentralised Coordination Approach for Multi-Robot Production Planning in Open Shared Factories}
\tnotemark[1,2]
\tnotetext[1]{This is the manuscript version accepted in the journal Robotics and Computer-Integrated Manufacturing. DOI: 10.1016/j.rcim.2020.102110}
\tnotetext[2]{\copyright 2020. This manuscript version is made available under the CC-BY-NC-ND 4.0 license http://creativecommons.org/licenses/by-nc-nd/4.0/}
%

\author[1]{Marin Lujak}[orcid=0000-0001-8565-9194]
\ead{marin.lujak@imt-lille-douai.fr}
\address[1]{IMT Lille Douai,
Universit\'e de Lille, CERI Numerique, 59000 Lille, France}

\author[2]{Alberto Fern\'andez}[orcid=0000-0002-8962-6856]
\ead{alberto.fernandez@urjc.es}
\address[2]{University Rey Juan Carlos,  28933 M\'ostoles, Madrid, Spain}

\author[3]{Eva Onaindia}[orcid=0000-0001-6931-8293]
\ead{onaindia@dsic.upv.es}
\address[3]{Valencian Research Institute for AI,  Universitat Polit\`ecnica de Val\`encia, 941,  46022 Valencia,  Spain}


\begin{abstract}
Open and shared manufacturing factories typically dispose of a limited number of industrial  robots and/or other production resources  that should be properly allocated to tasks in time for an effective and efficient system performance. In particular, we deal with the dynamic capacitated production planning problem with sequence independent setup costs where quantities of products to manufacture need to be determined at consecutive periods within a given time horizon and products can be anticipated or back-ordered related to the demand period. We consider a decentralised multi-agent variant of this problem in an open factory setting with multiple owners of robots as well as different owners of the items to be produced, both considered   self-interested and individually rational.
Existing solution approaches to the classic constrained lot-sizing  problem are  centralised exact methods that  require sharing of global knowledge of all the participants' private and sensitive information and are not applicable in the described multi-agent context. Therefore, we propose a computationally efficient decentralised approach based on the spillover effect that solves this NP-hard problem by distributing decisions in an intrinsically decentralised multi-agent system environment while protecting private and sensitive information. To the best of our knowledge, this is the first decentralised algorithm for the solution of the studied problem in intrinsically decentralised environments where production resources and/or products are owned by multiple stakeholders with possibly conflicting objectives. To show its efficiency, the performance of the Spillover Algorithm is benchmarked against state-of-the-art commercial solver CPLEX 12.8.
\end{abstract}

%
%
%
%


\maketitle

\section{Introduction}
\label{IntroOCOP}

The smart factory concept (\eg \cite{Shrouf2014,hozdic2015smart}) in Industry 4.0 revolution is opening up new ways of addressing the needs of sustainability and efficiency in the manufacturing industry of today's global economy.
In particular, we are seeing the emergence of shared factories (\eg \cite{Jiang2019}) in which the owners of production resources (robots) do not necessarily produce their own products. Indeed, they may offer their facilities and available production resources therein to various other manufacturers that manufacture their products in the same shared facility. Even more, multiple factories could be linked in a more flexible global virtual facility (\eg \cite{hao2018virtual}).

In this paper, we consider an (intrinsically decentralised)  shared factory scenario that requires decentralised methods for efficient allocation of robots (production resources)  of the same shared manufacturing plant to individually rational and self-concerned firms (users) that  use them to manufacture their products in a given time horizon. We consider open firms that may collaborate at times on common projects based on individual interest such that the factory's production capacity may vary from one period to another based on available shared resources. The requisite for  shared factories is that they follow the concept of an open firm, \ie  a common standard  for the production of components  and their interfaces is shared by distinct firms and each firm is open to all industry participants (\eg \cite{farrell1998vertical,arora2007open}).
An important aspect to consider  in this context is that the different firms participating in such manufacturing ecosystems are not willing to share  information or business strategy. Therefore, solutions in which a central coordinator receives all the information regarding resource availabilities and production demands are not applicable.

Production planning considers the best use of resources to satisfy the demand in a given planning time horizon. A production plan must meet conflicting objectives of guaranteeing service quality and minimizing production and inventory costs. To this end, the Constrained Lot-Sizing Problem (CLSP) answers the question of when and how much of each product (item) must be produced so that the overall costs are minimised (\eg ~\cite{buschkuhl2010dynamic,karimi2003capacitated}).

In this paper, we focus on deterministic, dynamic, single-level, multi-item CLSP with back orders and sequence independent setup costs, which we will refer to as MCLSP-BOSI (\eg \cite{Pochet06}). The concept of dynamic in MCLSP-BOSI means that the production demands vary through time;
single-level indicates that the end product or item is directly manufactured in one step from raw materials with no intermediate sub-assemblies and no item can be a predecessor of another item; multi-item refers to the existence of multiple product types and capacitated denotes a scarcity between the available limited manufacturing resources and the production demand.
Moreover, the presence of back orders implies that it is possible to satisfy the demand of the current period in future periods by paying shortage costs (back-ordering). We consider a lost-sales inventory model where a unit of a back-ordered product that cannot be produced at the period of its demand is requested nonetheless  for when resources become available again in some later period of the planning time horizon, while a backlogged unit is not produced at all and the demand is lost  resulting in lost sales (\eg  \cite{bijvank2011lost}). The sequence independent setup costs include the setup of robots and other production resources for producing an item at some period independently of its previous production dynamics (\eg ~\cite{Pochet06}). Contrary to   most of the related work, in our studied MCLSP-BOSI model, all costs are time-dependent.

The studied MCLSP-BOSI is a variant of the CLSP where production can be anticipated or delayed in relation to the product demands, resulting in time-dependent holding and back order costs, respectively. The inclusion of back orders is crucial in the contexts of high demands, since, otherwise, no feasible production plan would exist. Item back-ordering results in not producing (lost sales) if the overall demand is higher than the production capacity over a given time horizon.  The main challenge in back-ordering is to select lost sales (if any), the items to back-order,  the ones to produce at the time of demand release, and the  ones beforehand (\cite{drexl1997lot,jans2008modeling}). For an overview of the CLSP and its variations, we refer the reader to excellent descriptions in \cite{bitran1982computational,Pochet06,quadt2008capacitated}.

The MCLSP-BOSI is an NP-hard problem (\cite{chen1990analysis,bitran1982computational,dixon1981heuristic}) and has been mostly addressed by heuristic approximations that do not guarantee an optimal solution, but find a reasonably good solution in a moderate amount of computation time. In its classical mathematical form, MCLSP-BOSI is intrinsically centralised, i.e., a single decision-maker has the information of all the production resources and all the items at disposal. This formulation and related state-of-the-art solution approaches are not applicable in intrinsically decentralised contexts, such as open and shared factories.

With the aim of finding a production plan in intrinsically decentralised open and shared factories, in this paper, we mathematically formulate a decentralised variant of the MCLSP-BOSI problem.
As a solution approach to this problem, we propose the \textit{Spillover Algorithm}, a decentralised and heuristic   production planning method whose approximation scheme is based on the \emph{spillover effect}.
The spillover effect is defined as a situation that starts in one place and then begins to expand or has an effect elsewhere. In ecology, this term refers to the case where the available resources in one habitat cannot support an entire insect population, producing an ``overflow'', flooding adjacent habitats and exploiting their resources (\cite{rand2006spillover}). In economy, it is related to the interconnection of economies of different countries where a slight change in one country's economy leads to the change in the economy of others.

We leverage the spillover effect in  a multi-agent system  considering both the self-concerned owners of production resources and the self-concerned owners of the items' demands.
For a better explanation of the proposed multi-agent system, the spillover effect is modelled in a  metaphor of a liquid flow network with buffers, where the capacity of each buffer represents available production capacity at a  specific period.
Liquid agents assume the role of demand owners, with liquid volume proportional to the production demand.
Buffer agents, on the other hand, act in favour of  resource owners by allocating their production capacities to liquid agents. Liquid demand that cannot be allocated where requested, spills over through a tube towards other neighbouring buffers based on the rules that dictate how the spillover proceeds.
In the proposed Spillover algorithm, we assume that all agents are truthful (non-strategic), i.e. the information they provide is obtained from their true internal values and is not strategically modified to obtain better resource allocation.

To the best of our knowledge, this is the first decentralised heuristic approach that tackles the high computational complexity of the studied MCLSP-BOSI problem with time-dependent costs and is applicable in intrinsically decentralised shared and open factories.

This paper is organized as follows. We give an overview of the related work in Section \ref{RelatedWork}. In Section \ref{ProblemFormulation}, we present the background   of  the studied classic and centralised MCLSP-BOSI problem. In Section  \ref{MCLSP-BOSIFormulation}, we present the motivation and mathematical formulation of the decentralised MCLSP-BOSI problem  and discuss  our problem decomposition and decentralisation approach.
Our multi-agent architecture together with the Spillover Algorithm is presented in Section \ref{SpilloverAlgorithm}. Section \ref{experiment} presents the evaluation results of simulated experiments whose setup was taken from related work, while Section \ref{Conclusions} concludes the paper  and outlines future work.

\section{Related work}
\label{RelatedWork}

Due to its complexity, the literature on the classic (centralised) MCLSP-BOSI problem is scarce and can be classified into: (a) \textit{period-by-period} heuristics, which are special-purpose methods that work in the time horizon from its first to the last period in a single-pass construction algorithm; (b) \textit{item-by-item} heuristics, which iteratively schedule a set of non-scheduled items; and (c) \textit{improvement heuristics } that are  mathematical-programming-based heuristics that start with an initial (often infeasible) solution for the complete planning horizon usually found by uncapacitated lot sizing techniques, and then try to enforce feasibility conditions, by shifting lots from period to period at minimal extra cost. The aim of improvement heuristics is to maximise cost savings as long as no new infeasibilities are incurred (\eg ~\cite{Maes88,buschkuhl2010dynamic}).
However, these general heuristic approaches in many cases do not guarantee finding a feasible solution of the MCLSP-BOSI problem even if one exists (~\cite{millar1994lagrangian}).

~\cite{pochet1988lot} approach this problem by a shortest-path reformulation solved by integer programming and by a plant location reformulation for which they propose a cutting plane algorithm. Both approaches produce near optimal solutions to large problems with a quite significant computational effort. \cite{millar1994lagrangian} present two Lagrangian-based algorithms to solve the MCLSP-BOSI  based on Lagrangian relaxation and Lagrangian decomposition. However, their MCLSP-BOSI model is limited as it considers setup, holding, and back order costs while considering production costs fixed and constant throughout the planning horizon. They find a feasible primal solution  at each Lagrangian iteration and guarantee finding a feasible solution if one exists with valid lower bounds, thus, guaranteeing the quality of the primal solutions.

~\cite{KarimiGW06} present a tabu search heuristic method consisting of four parts that provide  an initial feasible solution: (1) a demand shifting rule, (2) lot size determination rules, (3) checking feasibility conditions and (4) setup carry over determination. The found feasible solution is improved by adopting the setup and setup carry over schedule and re-optimising it by solving a minimum-cost network flow problem. The found solution is used as a starting input to  a tabu search algorithm. An overview of metaheuristic approaches to dynamic lot sizing can be found in, \eg \cite{jans2007meta}.
Furthermore, there exist heuristic-based algorithms based on local optima (\eg ~\cite{Cheng01}), approaches that examine the Lagrangian relaxation and design heuristics to generate upper bounds within a subgradient optimisation procedure (\eg ~\cite{SuralDW09}), or a genetic algorithm for the multi-level version of this problem that uses fix-and-optimise heuristic and mathematical programming techniques (\eg ~\cite{ToledoOF13}).

\cite{zangwill1966deterministic} presents a CLSP model with backlogging where the demand must be satisfied within a given maximum delay while \cite{pochet1988lot} study uncapacitated lot sizing problem with backlogging and present the shortest path formulation and the cutting plane algorithm for this problem.
Furthermore, ~\cite{quadt2008capacitated} study a MCLSP-BO with setup times and propose a solution procedure  based on a novel aggregate model, which uses integer instead of binary variables. The model is embedded in a period-by-period heuristic and is solved to optimality or near-optimality in each iteration using standard procedures
(CPLEX). A subsequent scheduling routine loads and sequences the products on the parallel machines. Six versions of the heuristic were presented and tested in an extensive computational study showing that the heuristics outperforms the direct problem implementation as well as the lot-for-lot method.

~\cite{goren2018fix} integrate several problem-specific
heuristics  with fix-and-optimise (FOPT) heuristic. FOPT heuristic is a
MIP-based heuristic in which a sequence of MIP models is solved over all real-valued decision
variables and a subset of binary variables. Time and product decomposition schemes are
used to decompose the problem. Eight different heuristic approaches are
obtained and their performances are shown to be promising in simulations.

From the decision distribution perspective, the above mentioned solution approaches  solve the MCLSP-BOSI problem by a single decision-maker having total control over and information of the production process and its elements.
 Since the MCLSP-BOSI problem is highly computationally expensive, decomposition techniques may be used for simplifying difficult constraints. They were applied by \cite{millar1994lagrangian} who used Lagrangian relaxation of  setup constraints, while  \cite{thizy1985lagrangean} studied the multi-item CLSP problem without back orders and used Lagrangian relaxation of capacity constraints to decompose the problem into     single item uncapacitated lot sizing sub-problems solvable by the Wagner-Whitin algorithm (see, \cite{wagner1958dynamic}) and its improvements (\eg \cite{aggarwal1993improved,brahimi2017single}.  Moreover, \cite{lozano1991primal}, \cite{diaby1992lagrangean}, and \cite{hindi1995computationally} are some other works using Lagrangian relaxation. A broad survey of single-item  lot-sizing problems can be found in \cite{brahimi2017single}.
All these Lagrangian relaxation and decomposition methods focus on solving the CLSP problem centrally by one decision maker on a single processor.
State-of-the-art centralised heuristic solution approaches and related surveys can be found in, \eg \cite{karimi2003capacitated}, \cite{KarimiGW06}, \cite{quadt2009capacitated}, \cite{goren2018fix}, and \cite{jans2008modeling}.

While the MCLSP-BOSI state-of-the-art solution approaches  improve solution quality in respect to previous work, they are all centralised approaches that can only be applied for a single decision maker.
\cite{giordani2009decentralized} and \cite{giordani2013distributed} deal with distributed multi-agent coordination in the constrained lot-sizing context. Both works assume the existence of a single robot (production resource) owner agent  responsible for achieving globally efficient robot allocation by interacting with  product (item) agents through an auction. The problem decomposition is done  to gain computational efficiency since item agents can compute their bids in parallel. The allocation  of limited production resources   is done through the interaction between competing item agents and a robot owner (a single autonomous agent) having available all the global information.
However, these approaches are not adapted to intrinsically decentralised scenarios as is the case of competing shared and open factories since the resource allocation decisions  are still made by a single decision maker (robot owner) based on the bids of the item agents that are computed synchronously.

In shared and open factories, both the interests of  rational self-concerned owners of the production resources as well as the interests of  self-interested rational owners of the items' demands must be considered with minimal exposure of their  sensitive private information.
Distributed constraint programming approaches are computationally expensive (\eg \cite{faltings2006distributed}) and are, therefore, not applicable in such large systems producing multitudes of items.

Cloud manufacturing is an interesting paradigm that integrates distributed resources and capabilities, encapsulating them into services that are managed in a centralized way \cite{XU201275}. Important issues with cloud manufacturing are how to schedule multiple manufacturing tasks to achieve optimal system performance or how to manage the resources precisely and timely. Workload-based multi-task scheduling methods \cite{LIU20173} or approaches of resource socialization \cite{QIAN2020101912} have been proposed to address the limitations inherent of the centralized structure of cloud platforms. Unlike cloud manufacturing approaches, we advocate for a decentralized approach to the manufacturing planning problem that achieves a good quality solution while protecting sensitive private information of the stakeholders.

In this paper we present a first solution approach for the decentralised variant of the MCLSP-BOSI problem; a sketchy outline of this approach can be seen in \cite{Lujak0O20}.

\section{Overview of the studied MCLSP-BOSI problem}
\label{ProblemFormulation}

The objective of the MCLSP-BOSI is to find  a  production schedule for a set $I$ of different types of items that  minimizes  the total back order, holding inventory, production,  and setup costs over a given finite time horizon $T$ subject to demand and capacity constraints. This problem belongs to the class of deterministic dynamic lot-sizing problems well known in the inventory management literature (\eg   \cite{Pochet06,buschkuhl2010dynamic,jans2008modeling,karimi2003capacitated,millar1994lagrangian,quadt2008capacitated}).
In the following, we give the formulation of the baseline problem.

Before the beginning of the time horizon $T$,   the customer demand arrives for each item $i\in I$ and each period $t\in T$ of the time horizon and is received as incoming order.
Items $i \in I$, with given resource requirements $r_i$, are produced at the beginning of each period by robots (manufacturing resources) of a given limited and time-varying production capacity $R_t$ known in advance for the whole given time horizon $T=\{1, \ldots |T|\}$. Open and shared factories may collaborate at times based on individual interest such that the factory's production capacity may vary through time based on available shared resources.
We assume that a time-varying and sequence independent setup cost $s_{it}$ that includes the setup of the production resources for the production of item $i$ occurs  if item $i$ is produced at period $t$ and a linear production cost $c_{it}$ is added for each unit produced of item $i$ at the same period. Setup costs usually include the preparation of the robots, \eg  changing or cleaning a tool, expediting, quality acceptance, labour costs, and opportunity costs since setting up the robot consumes time.
Moreover, demand $d_i$ for each item $i\in I$ can be anticipated or delayed  in respect to the period in which it is requested. If anticipated, it is at the expense of  a linear  holding cost $h_{it}$ for each unit of item $i$ held in inventory per unit period and if delayed, a linear back order cost  $b_{it}$ is accrue for every unit back ordered per unit period.  
The  inventory level and back orders are measured at the end of each period.

Each item demand $\mathbf{d_i}=\{d_{i1}, \ldots, d_{i,|T|}\}$ is associated with inventory level $x_{it}$ at the end of period $t \in T$ that can be positive (if a stock of completed  items  $i$ is present in the buffer at the end of  period $t$), zero (no stock and no back order), or negative (if a back order of demands for item $i$ is in the queue at the end of period $t$). The inventory level is increased by production quantity $u_{it}$  and reduced by demand $d_{it}$ at each period $t$.

Using a standard notation, we denote $x^+_{it}:=\max\{x_{it},0\}$ as the stock level, and $x^-_{it}:=\max\{-x_{it},0\}$ as the back order level at the end of  period $t$.
Stock $x^+_{it}$ available at the end of period $t$ for product $i$ corresponds to the amount of product $i$ that is physically stored on stock and is available for the demand fulfilment. Notice that, for each  period $t$ and item $i$, $x^+_{it}\cdot x^-_{it}=0$, \ie   positive back order value and storage value cannot coexist at the same time since if there is a back order of demand, it will be first replenished by the items we have in the storage and then, we will produce the rest of the demand.

The problem is to decide when and how much to produce in each period of time horizon $T$ so that all demand is satisfied at minimum cost.
The mathematical program $(P)$ of the studied MCLSP-BOSI problem  is given in the following; related
sets, indices, parameters, and decision variables are summarised
in Table \ref{Table:SymbolsModel}.

\begin{table}
    \centering
    \caption{List of symbols used in the MCLSP-BOSI problem modelling}
    \label{Table:SymbolsModel}
\begin{tabular}{ll} \toprule
\multicolumn{2}{c}{Sets and indices} \\
\midrule
$T$ & planning time horizon, set of consecutive periods $t \in T$ of equal length\\
$I$& set of items $i \in I$ \\
  \midrule
  \multicolumn{2}{c}{Parameters} \\
  \midrule
  $d_{it}$ & demand for item $i$ at period $t$\\
	 $r_i$ & resource   requirement  for production of a unit of item $i$\\
	$c_{it}$ & unitary production cost for item $i$ at period $t$ \\
   $h_{it}$& unitary holding cost for item $i$ at period $t$\\
  $b_{it}$ & unitary back order cost for item  $i$ at period $t$\\
   $s_{it} $ &  fixed setup cost for item  $i$ at period $t$\\
   $R_t$ &  production  capacity at period $t$\\
  \midrule
  \multicolumn{2}{c}{Decision variables} \\
  \midrule
$u_{it}$ & number of items  $i$  produced at period $t$\\
$x_{it}^+, x_{it}^-$ & storage and back order inventory level of item $i$ at the end of period $t$\\
$y_{it}$ & equals 1 if  item $i$ is produced at period $t$, 0 otherwise. \\

    \midrule
  \multicolumn{2}{c}{Objective function} \\
  \midrule
  $z(\mathbf{u})$ &  generalized cost objective function of solution $\mathbf{u}_{it}$, $i \in I$, $t \in T$ \\
  \bottomrule\\
  \end{tabular}
\end{table}
%

\indent ($P$):\\
\begin{equation}\label{minPiG}
z(\mathbf{u})= \min \sum_{i \in I, t \in T}\big(h_{it} x^+_{it} + b_{it} x^-_{it} + c_{it} u_{it}+ s_{it} y_{it}\big)+ \sum_{i \in I}\big(b_{i,|T|+1}x_{i,|T|+1}^- + h_{i,|T|+1}x_{i,|T|+1}^+\big)
\end{equation}

\indent subject to:
\begin{equation}
\label{balanceG}
x^+_{it} - x^-_{it} = x^+_{i, t-1} - x^-_{i, t-1} + u_{it}-d_{it},\;\; \forall i \in I,\; \forall t \in T
\end{equation}

\begin{equation}
\label{balanceGT1}
x^-_{i,|T|+1}-x^+_{i,|T|+1} = x^-_{i,|T|} - x^+_{i,|T|} ,\;\; \forall i \in I
\end{equation}

\begin{equation}
\label{ulimitG}
\sum_{i \in I} u_{it} r_{i} \leq  R_t , \,\,\, \forall t \in T
\end{equation}

\begin{equation}
\label{setupCostsG}
u_{it} \leq y_{it}\sum_{k \in T}d_{ik}, \,\,\,\forall  i\in I, \;\forall\;t \in T
\end{equation}

\begin{equation}
\label{variableG}
u_{it}  \in \mathbb{Z}_{\geq 0},  \,\,\,\forall i \in I, \; \forall t \in T
\end{equation}
\begin{equation}
\label{variableintegerG}
y_{it} \in \{0,1\}, \,\,\,\forall  i \in I, \; \forall t \in T
\end{equation}
\begin{equation}
\label{variableGT1P}
x^+_{it}, x^-_{it} \in \mathbb{Z}_{\geq 0},  \,\,\,\forall i \in I, \; \forall  t \in   \{1, \ldots, |T|+1\}
\end{equation}

The values of $x^+_{i0}$ and $x^-_{i0}$ represent the initial conditions, \ie the stock and back order level of item $i$ at the beginning of the planning time horizon.

For this model  to accommodate not only back orders but also lost sales (backlog)  and surplus (unsold) stock after the end of the planning time horizon $T$,  in addition to the sum of the sustained   costs in a given time horizon, in  (\ref{minPiG}), there is also an additional backlog (lost sales) cost   \ie $\sum_{i\in I}b_{i,|T|+1}x_{i,|T|+1}^-$ and surplus (unsold) stock  cost   \ie $\sum_{i\in I}h_{i,|T|+1}x_{i,|T|+1}^+$.
Generally, the decisions on surplus stock $x_{i,|T|+1}^+$ and  backlog $x_{i,|T|+1}^-$ (demand that is not produced in the time horizon) will depend on the value of surplus stock cost  $h_{i,|T|+1}$ and
 backlog cost $b_{i,|T|+1}$ and their relation to production $c_{it}$, holding $h_{it}$, and back order $b_{it}$ costs, as well as the relation between overall demand and the overall production capacity throughout  time horizon $T$.

Constraints (\ref{balanceG}) are  flow-balance constraints among product demand $d_{it}$, stock level $x^+_{it}$, back order level $x^-_{it}$, and production level $u_{it}$, for each period $t \in T$. Constraints (\ref{balanceGT1}) are the flow-balance constraints for the end of the time horizon $|T|+1$, which is the reason why the domain of index $t$ of $x_{it}^-$ and $x_{it}^+$ in (\ref{variableGT1P}) runs  in interval $\{1, \ldots, |T|+1\}$.
Without loss of generality and assuming that for each period $t$ and  item $i$, the values of $h_{it}$ and $b_{it}$ are non-negative (zero or positive), the constraint $x^+_{it}\cdot x^-_{it}=0$ is implicitly satisfied by the optimal solution, and, hence, omitted in the formulation.

Constraints (\ref{ulimitG}) limit the overall resource usage for the production of all the items not to be greater than the production capacity  at period $t$. Note that production capacity $R_t$ of each period $t \in T$ may vary from one period to another.
Contrary to the classic MCLSP-BOSI model, our studied model does not have a strong assumption that the overall demand in the time horizon is lower than the overall production capacity.
Here,  the following constraints on   resource   requirements for production of a unit of each item $i\in I$ hold:
\begin{equation}\label{largeLiquids}
r_i\leq \max_{t \in T}\{R_t\}, \;\; \forall i \in I,
\end{equation}
\begin{equation}\label{smallContainers}
R_t \geq \min_{i \in I}\{r_i\}, \;\; \forall t \in T.
\end{equation}
Items $i$ that violate assumption (\ref{largeLiquids}) and periods $t$ that violate assumption (\ref{smallContainers}) can be removed.
By constraints (\ref{setupCostsG}), we model independent setups and the fact that if the production is launched for item $i$ in period $t$, the quantity produced should not be larger than the overall demand for item $i$ in a given time horizon.  Constraints (\ref{variableG}) and (\ref{variableGT1P}) are nonnegativity and integrality constraints on production, back order and storage  decision variables, while constraints (\ref{variableintegerG})  limit  the setup decision variables to binary values, \ie  the fixed setup cost is accrue if there is (any) production at a period.

The solution to problem $P$ is a production plan composed of  a number of items $u_{it}$  to produce for each item type $i \in I$ in response to  demand $d_{it}$ in each period $t \in T$ of a given time horizon.

\paragraph{Complexity of the MCLSP-BOSI problem.}
The single-item capacitated lot-sizing problem is NP-hard even for many special cases (\eg  \cite{florian1980deterministic,bitran1982computational}.
\cite{chen1990analysis}) proved that multi-item capacitated lot-sizing problem (MCLSP) with setup times is strongly NP-hard.
Additionally, \cite{bitran1982computational} present the multiple items capacitated lot size model with independent setups without back orders and resource capacities that change through time. Contrary to our model that assumes non-negative integer values for $u_{it}, x^+_{it}$, and $x^-_{it}$, the model in \cite{bitran1982computational} contains no integrality constraints for these variables.
The introduction of an integrality constraint leads to a generally NP-complete integer programming problem.

\section{Formulation of the decentralised MCLSP-BOSI problem}

Problem $P$ (\ref{minPiG})--(\ref{variableGT1P}) is a classic and centralised mathematical formulation of the MCLSP-BOSI problem.
The decentralised variant of the MCLSP-BOSI problem should consider that both the providers of capacitated and homogeneous robots  and their customers (the owners of the produced items) are interested in the transformation of heterogeneous raw materials into heterogeneous final products and thereby both of them should be considered active participants in the production process; no one is willing to disclose its complete information but will share a part of it if this facilitates achieving its local objective.
Therefore, they must negotiate resource allocation while exchanging relevant (possibly obsolete) information.


\label{MCLSP-BOSIFormulation}
In the following,  we propose the formulation of the decentralised MCLSP-BOSI problem.

First, we create a so-called Lagrangian relaxation (\eg \cite{lemarechal2001lagrangian,fisher1981lagrangian}) of problem $(P)$, which we name $L(P,\Lambda)$, for which we assume to be  tractable. This new modelling allows us to decompose the original problem $P$ into item-period pair $(i,t)$ subproblems, where $i \in I$ and $t \in T$ (section \ref{ItemPeriodModelling}). The decomposition based on item-period  $(i,t)$ pairs allows for the agentification of the demand of product type $i$ at period $t$ and decentralisation of  problem $(P)$.


The Lagrangian relaxation is achieved  by turning the complicating global capacity constraints (\ref{ulimitG}) into constraints that can be violated at   price $\Lambda=\{\lambda_t |  t \in T\}$, while keeping the remaining (easy) constraints  for each  product $i \in I$ and period   $t \in T$.  We dualize  the complicating constraints (\ref{ulimitG}), \ie we  drop them while adding their slacks   to the objective function with weights $\Lambda$ (vector of dual variables for constraints (\ref{ulimitG})-Lagrangian multipliers) and thus create the \textit{Lagrangian relaxation} $L(P,\Lambda)$ of $P$.

 Here, we let $\Lambda$ be the (current) set of nonnegative Lagrange multipliers (resource conflict prices).

The solution of the Lagrangian problem provides a lower bound, while the upper bound is
obtained by first fixing the setup variables given by the dual solution and secondly, obtaining the solution from the resulting transportation problem.

Then the  Lagrangian relaxation of the MCLSP-BOSI problem can be mathematically modelled as follows:\\
\indent ($L(P,\Lambda)$):\\
\begin{equation}\label{minPiGLagrange}
z^*(\mathbf{u})= \min \sum_{i \in I, t \in T}\big(h_{it} x^+_{it} + b_{it} x^-_{it} + c_{it} u_{it}+ s_{it} y_{it}\big)+ \sum_{i \in I}\big(b_{i,|T|+1}x_{i,|T|+1}^- + h_{i,|T|+1}x_{i,|T|+1}^+\big)
+\sum_{i \in I, t \in T}\lambda_t\big(u_{it} r_{i} -R_t\big)
\end{equation}

\indent subject to:
\begin{equation}
\label{balanceGLagrange}
x^+_{it} - x^-_{it} = x^+_{i, t-1} - x^-_{i, t-1} + u_{it}-d_{it},\;\; \forall i \in I,\; \forall t \in T
\end{equation}

\begin{equation}
\label{balanceGT1Lagrange}
x^+_{i,|T+1|} - x^-_{i,|T+1|} = x^+_{i,|T|} - x^-_{i,|T|} ,\;\; \forall i \in I
\end{equation}


\begin{equation}
\label{setupCostsGLagrange}
u_{it} \leq y_{it}\sum_{k \in T}d_{ik}, \,\,\,\forall  i\in I, \;\forall\;t \in T
\end{equation}
\begin{equation}
\label{variableGLagrange}
u_{it}  \in \mathbb{Z}_{\geq 0},  \,\,\,\forall i \in I, \; \forall t \in T
\end{equation}
\begin{equation}
\label{variableintegerGLagrange}
y_{it} \in \{0,1\}, \,\,\,\forall  i \in I, \; \forall t \in T
\end{equation}
\begin{equation}
\label{variableGT1PLagrange}
x^+_{it}, x^-_{it} \in \mathbb{Z}_{\geq 0},  \,\,\,\forall i \in I, \; \forall  t \in   \{1, \ldots, |T|+1\}
\end{equation}
where the variable, parameter, and constraint descriptions remain the same as in $(P)$ except for constraints (\ref{ulimitG}) that are relaxed in (\ref{minPiGLagrange}).

\subsection{Item-period agent modelling}
\label{ItemPeriodModelling}

We decompose the overall production planning problem to subproblems solvable by item-period $(i,t)$ pair agents.
Given a vector $\Lambda$,  we substitute indices $a=(i,t)$.
%
The mathematical program of the problem $L(P_A,\Lambda)$ reformulated for $(i,t)$ pair agent $a \in A$ is given in the following; related sets, indices, parameters, and decision variables are summarised in Table \ref{Table:SymbolsModelItemPeriod}.

\begin{table}[h]
    \centering
    \caption{List of symbols used in the item-period pair agent modelling}
    \label{Table:SymbolsModelItemPeriod}
  \begin{tabular}{ll}
  \toprule
  \multicolumn{2}{c}{Sets and indices} \\
  \midrule
  $T$ & planning time horizon, set of consecutive periods $k,t \in T$ of equal length\\
  $A$ & set of pairs $a=(i,t)$,   where $i \in I$ and $t \in T$ whose elements $a \in A$  have \\
      & positive demand value, \ie $d_{it}>0$ \\

  \midrule
  \multicolumn{2}{c}{Parameters} \\
  \midrule
 $d_{a}$ & demand for pair $a=(i,t)$, \ie $d_{a}=d_{it}$ \\
  $d_{ak}$ & demand for pair $a=(i,t)$ at period $k$, where $d_{ak}=d_{it}$ when $k=t$, and $d_{ak}=0$ otherwise\\
	 $r_a$ & resource requirement for production of a unit of pair $a$ (item $i$ at period $t$) \\
	$c_{a}$ & unitary production cost for pair $a$\\
   $h_{ak}$& unitary holding cost for pair $a$ at period $k$\\
  $b_{ak}$ & unitary back order cost for pair  $a$ at period $k$\\
   $s_{ak} $ &  fixed setup cost for pair  $a$ at period $k$\\
   $R_k$ & production capacity at period $k$\\

  \midrule
  \multicolumn{2}{c}{Decision variables} \\
  \midrule
  $u_a^k$	& number of items $i$ demanded at time $t$ ($a=(i,t)$) that are produced  at time $k$. \\
     & Note that we  use $k$ as superindex to avoid confusion  between \\
  & $u_a^k$ and $u_{it}$ from $(P)$, which are not the same; $u_a^k=u_{itk}$\\

$x_{ak}^+, x_{ak}^-$ & holding and back order buffer content of pair $a$ at the beginning of period $k$\\
$y_{ak}$ & equals 1 if  pair $a$ is produced at period $k$, 0 otherwise. \\

  $\mathbf{u}_a$ & vector $\{ u_a^k  | \forall a \in A, k \in T \}$\\ 
   \midrule
  \multicolumn{2}{c}{Objective function} \\
  \midrule

  $z_a(\mathbf{u}_a)$ & generalized cost objective function of solution $\mathbf{u}_a$, $a \in A$ \\

  \bottomrule\\
  \end{tabular}
  \end{table}

\indent ($L(P_A,\Lambda)$):
\begin{multline}\label{minPiL}
z^*(\mathbf{u})=\min  \sum_{a \in A} \sum_{k \in T} \Big(h_{ak} x^+_{ak} + b_{ak} x^-_{ak} + c_a u_a^k+ s_{ak} y_{ak}\Big)\\
+ \sum_{a\in A}\big(b_{a,|T|+1} x_{a,|T|+1}^- +h_{a,|T|+1}x_{a,|T|+1}^+ \big)
+\sum_{k \in T}\lambda_k\big(\sum_{a \in A}u_a^k r_{a} -R_k\big)
\end{multline}
\indent subject to:
\begin{equation}
\label{balanceL}
x^+_{ak} - x^-_{ak} = x^+_{a, k-1} - x^-_{a, k-1} + u_a^k-d_{ak},\;\; \forall a \in A,\forall k \in T
\end{equation}

\begin{equation}
\label{balanceGT1Lambda}
x^+_{a,|T+1|}- x^-_{a,|T+1|} = x^+_{a,|T|} - x^-_{a,|T|} ,\;\; \forall a \in A
\end{equation}

\begin{equation}
\label{setupCostsL}
 u_a^k \leq   d_{a} y_{ak}, \,\,\,\forall a \in A, \forall  k \in T
\end{equation}
\begin{equation}
\label{variableL}
u_a^k \in \mathbb{Z}_{\geq 0},  \,\,\, \forall a \in A, \forall  k \in T
\end{equation}
\begin{equation}
\label{variableintegerL}
y_{ak} \in \{0,1\}, \,\,\,\forall a \in A, \forall  k \in T
\end{equation}
\begin{equation}
\label{variableGT1Lambda}
x^+_{ak}, x^-_{ak}  \in \mathbb{Z}_{\geq 0},  \,\,\,\forall a \in A, \; k \in T\cup \{|T|+1\},\\
\end{equation}
 where $\mathbf{u} = \{u_a^k | a \in A$ and $k \in T\}$, for all $a \in A$.
Similarly to problem (\textit{P}), the values of $x^+_{a0}$ and $x^-_{a0}$ represent the initial conditions, \ie the stock and back order level of pair $a=(i,t)$ at the beginning of the planning time horizon.
%
Constraints (\ref{balanceL}) are  flow-balance constraints among stock level $x^+_{ak}$, back order level $x^-_{ak}$, production level $u_a^k$, and product demand $d_{ak}$ for each period in time horizon $k \in T$. Note that $d_{ak}=d_a$ when $k=t$, \ie  at the period at which demand $a$ is released, and $d_{ak}=0$ for all other periods  $k \in T \backslash \{t\}$.

%
Constraints (\ref{balanceGT1Lambda}) model the backlog and surplus stock after the end of a given time horizon, at period $|T|+1$.
By constraints (\ref{setupCostsL}), we model the fact that if the production is launched in period $k$, the quantity produced should not be larger than the demand $d_a$ for pair $a=(i,t)$.
Constraints (\ref{variableL}) are nonnegativity and integrality constraints on production $u_a^k$ decision variables; constraints (\ref{variableintegerL})  limit  setup variables $y_{ak}$ to binary values and constraints (\ref{variableGT1Lambda}) model the domain of storage $x^+_{ak}$ and back order $x^-_{ak}$ variables to $\{ 1, \ldots, |T|+1\}$.

Since the formulation $L(P_A,\Lambda)$ is separable for a given set of multipliers, it is  possible to extract a local optimisation subproblem $P_a$ addressed by each $(i,t)$ pair agent $a\in A$ that includes its production quantities $u_a^k$, setup decisions $y_{ak}$,  holding  $x^+_{ak}$ and back order $x^-_{ak}$ levels  for each  period $k\in T$, as well as backlog $x^-_{a,|T|+1}$ and surplus stock $x^+_{a,|T|+1}$ after the end of planning time horizon $T$.

In the following, we present  the problem $P_a$ to be solved by each $(i,t)$ pair agent $a \in A$.\\

\indent ($P_{a}$):
\begin{equation}\label{minPi}
z_a(\mathbf{u_a}) =  \min\sum_{k \in T} \Big(h_{ak} x^+_{ak} + b_{ak} x^-_{ak} + s_{ak} y_{ak}+ c_a u_a^k\Big)
+ b_{a,|T|+1}x_{a,|T|+1}^- + h_{a,|T|+1}x_{a,|T|+1}^+
 +\sum_{k \in T} \lambda_k r_{a} u_a^k
\end{equation}
\indent subject to:

\begin{equation}
\label{balance}
x^+_{ak} - x^-_{ak} = x^+_{a, k-1} - x^-_{a, k-1} + u_a^k-d_{ak},\;\; \forall k \in T
\end{equation}

\begin{equation}
\label{balanceGT1aLambda}
x^+_{a,|T+1|}-x^-_{a,|T+1|} = x^+_{a,|T|} - x^-_{a,|T|}
\end{equation}

\begin{equation}
\label{ulimitInd}
u_a^k r_{a} \leq  R_k , \,\,\, \forall k \in T
\end{equation}
\begin{equation}
\label{setupCosts}
 u_a^k \leq   d_{a} y_{ak}, \,\,\,\forall  k \in T
\end{equation}
\begin{equation}
\label{variable}
u_a^k \in \mathbb{Z}_{\geq 0},  \,\,\,\forall  k \in T
\end{equation}
\begin{equation}
\label{variableinteger}
y_{ak} \in \{0,1\}, \,\,\,\forall  k \in T,\\
\end{equation}
\begin{equation}
\label{variableGT1}
x^+_{ak}, x^-_{ak}  \in \mathbb{Z}_{\geq 0},  \,\,\,\forall a \in A, \; k \in T\cup \{|T|+1\},\\
\end{equation}

where $\mathbf{u_a} = \{u_a^k |  k \in T\}$ and $\lambda_k$ is Lagrangian multiplier (conflict price) for resources available at period $k$. 
Here, except for the constraints already explained previously, additional constraints (\ref{ulimitInd}) limit the overall resource usage for the production of agent $a$ to be not greater than the amount of resources available at each period $k$. In this way, each $(i,t)$ pair agent resolves its local optimisation problem having available only its local variables and without the need to communicate with other agents to know their demands or production decisions.

The decentralised MCLSP-BOSI problem allows for multiple robot owner agents and multiple competing item-period pair agents requesting the production of the same item type at different periods, and  asynchrony in decision making.
For each $(i,t)$ pair agent $a \in A$, the solution of problem $P_a$ is a local production plan  $u_{a}=\{u_a^k| k=1, \ldots, |T|\}$
(in response to production demand $d_{a}$ and  the (current) set of nonnegative Lagrange multipliers \ie resource conflict prices $\lambda_k$) with related setup $y_{ak}$, storage $x_{ak}^+$, and back order $x_{ak}^-$ decisions in each period $k \in T \cup \{|T|+1\}$.
Plan $u_{a}=\{u_a^k| k=1, \ldots, |T|\}$  might not be globally feasible (\ie  not complying with constraints (\ref{ulimitG})) and thus should be negotiated with  other   agents $a \in A$.
Therefore, item-period pair agents must negotiate resource allocation while exchanging relevant (possibly obsolete) information. 
Resource allocation here should be achieved by the means of a decentralised protocol where fairness  plays a major role.  

%

%

\section{Liquid flow network model and the Spillover Algorithm}\label{SpilloverAlgorithm}
In this section, we propose  a decentralised multi-agent based coordination model  and the Spillover Algorithm for  the MCLSP-BOSI problem. In Table \ref{Table:symbolsSpillover}, we present symbols used in the Spillover Algorithm in addition to the ones already presented in Table \ref{Table:SymbolsModelItemPeriod}. We describe these symbols in the following.

\begin{table}
    \centering
    \caption{Additional symbols in the Spillover Algorithm}
    \label{Table:symbolsSpillover}

  \begin{tabular}{ll} \toprule
    $A$		& set of liquid agents $a \in A$, where $a = (i,t)$, $i \in I$ and $t \in T$ \\

  $\delta^+_{ak}$ $\delta^-_{ak}$ & valves controlling flow of liquid agent $a$ to posterior and anterior periods \\
  $\rho_{ak}$ & number of resources requested by liquid agent $a$ from buffer agent $k \in T$	\\

  $\mathbf{u}_k$ & vector $\{ u_a^k  | \forall a \in A \}$ of production accommodated by buffer $k$ \\
  $Z_a(\mathbf{u}_a)$ & approximated cost found by Spillover Algorithm solution $\mathbf{u}_a$, $a \in A$ \\

  \bottomrule
  \end{tabular}
\end{table}

We model each pair $(i,t)$, where $i \in I$ and $t \in T$ as a rational self-concerned   agent  $a \in A$ responsible for finding a production plan $u_{a}=\{u_{ak}| k=1, \ldots, |T|\}$ for the demand $d_a=d_{it}$, where  $A$ is a set of pairs $(i,t)$ of cardinality $|A|=|I|\cdot|T|$.
Since the heuristics that will be proposed in the following imitates the behaviour of liquid flows in tube networks (pipelines) with buffers, we call each agent $a\in A$ a \textit{liquid agent}.

\begin{figure}
\begin{center}
\includegraphics[width=0.6\columnwidth]{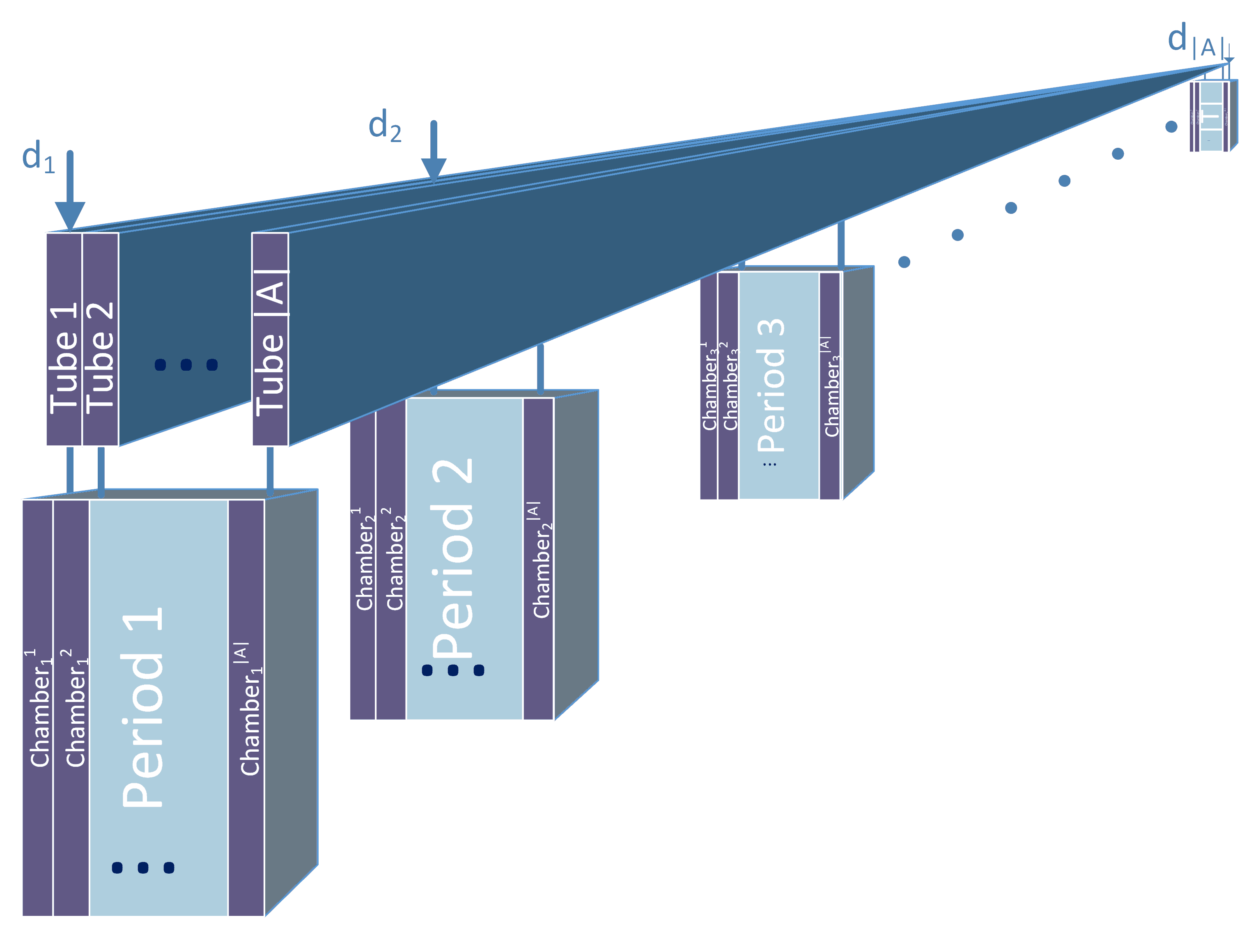}
\caption{Liquid flow network  model with $|T|$ buffers, one per each period; each buffer with $|A|$ impermeable chambers of interdependent volumes, one chamber per each liquid agent $a=(i,t)$}
\label{TimeHorizon}
\end{center}
\end{figure}

We model each period $k \in T$ as a limited resource allocation agent that we call a  \textit{buffer agent},  responsible for the allocation of its production capacity $R_k$ that is proportional to the volume of the buffer (inventory level), Figure \ref{TimeHorizon}. We assume that available shared resources in period $k \in T$ are owned by a single resource owner.

\begin{figure}[ht]
\begin{center}
\includegraphics[width=0.65\columnwidth]{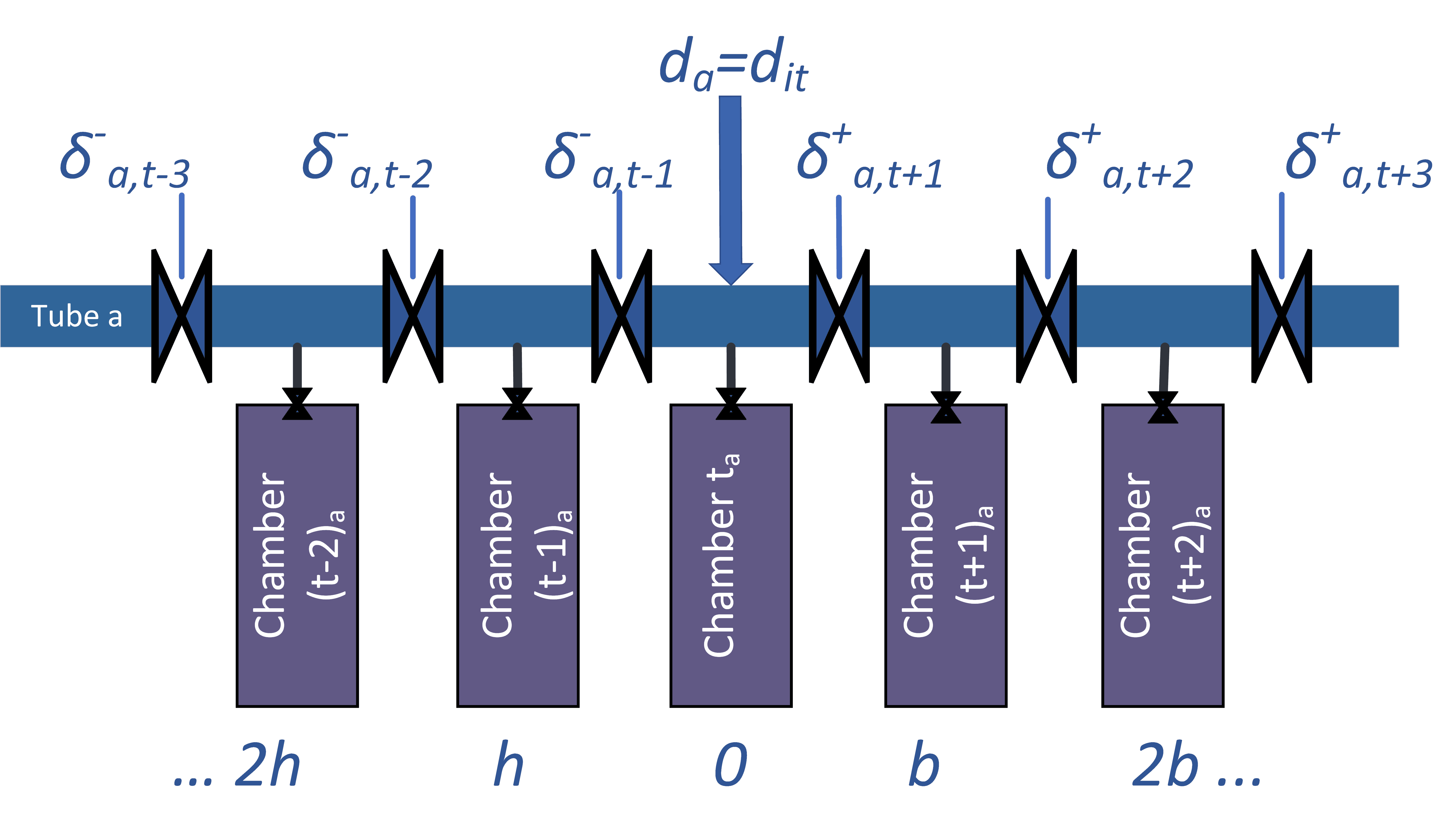}
\caption{Side view of a generic tube $a$ with related buffers' chambers in Figure \ref{TimeHorizon}}
\label{accumulatingCosts}
\end{center}
\end{figure}

Initially, each liquid agent $a$ releases liquid volume proportional to its demand $d_a$ into the horizontal tube above buffer $t$, Figure \ref{TimeHorizon}.
This volume, while in the horizontal tube,  is awaiting allocation (transfer) to one or more buffers.
A liquid agent $a$ negotiates producing in time, postponing   or advancing the production of its demand  with the buffers representing present, later  or earlier periods with respect to buffer $t$  by means of  two types of horizontal tube valves: valves  $\delta^{+}_{ak}$, that control the flow of liquid (demand) of  agent $a$ in the horizontal tube to the posterior periods  $k \in [t+1, \ldots, |T|]$  and  valves  $\delta^-_{ak}$, $k \in [t-1, \ldots, 1]$,  that control the flow of liquid $(i,t)$ to the previous periods $k \in [t-1, \ldots, 1]$ with respect to period $t$, as seen in Figure \ref{accumulatingCosts}.

The volume of each buffer  $k \in T$  is composed of $|I|\cdot |T|$ (initially empty) impermeable chambers, one chamber per each liquid agent $a=(i,t)$.
Each  chamber   is attached from above to  a horizontal tube exclusively dedicated to liquid $(i,t)$ through a valve controlled by the buffer agent, Figure \ref{accumulatingCosts}.
Buffer $k$ is modelled as a rational collaborative agent  that controls the valves regulating the flow  between  the horizontal tube of each liquid agent $a \in A$ that requests  allocation at period $k$ and buffer $k$.
So even if liquid $a$ is present above buffer $k$, it does not flow into the buffer if the valve of chamber $k_a$ of buffer $k$  does not open. The valve will open to let the inflow to chamber $k_a$ of the  liquid volume  corresponding to production $u_a^k$.
Initially, all buffers' valves and all liquids' horizontal tube valves   are closed. The volume of liquid $a = (i,t)$ equal to demand $d_{a}$ enters into the tube network above buffer $k=t$ ($t$ being the  period at which the demand is released). Valves $\delta_{ak}^{+}$ and $\delta_{ak}^-$ for every $k \in T$ are all closed  and do not allow the liquid flow in the horizontal tube, all liquid volume being concentrated between the valves $\delta_{a,t+1}^{+}$ and $\delta_{a, t-1}^-$, Figure \ref{accumulatingCosts}. The order of openings of these valves will control the direction of the flow  of the unallocated liquid demand in the horizontal tube. This order is influenced by the relation between accumulated back order and holding costs for the liquid and follows linear increase in their values as will be further explained.
Liquid agent $a=(i,t)$ can direct the flow of its demand from its dedicated tube $(i,t)$  into the chamber $k_a$ of  buffer $k \in T$ if and only if  all the control valves of liquid agent $a$ from period $t$ towards period $k$ are open  and the  valve of  chamber $k_a$ of buffer $k$ is open, Figure \ref{accumulatingCosts}.
The volume of flow from  tube  $(i,t)$  into buffer $k$ is equal to the production $u_a^k$. This value is inversely proportional to the liquid volumes of other chambers in the same buffer since their overall sum is limited from above by the buffer   capacity, i.e., $\sum_{a \in A}u_a^k \leq R_k$. 
When two or more liquid agents demand more resources than the production capacity $R_k$, they compete for them. Since no central resource allocation entity exists, each liquid agent needs to negotiate its production with buffer agents through   a negotiation mechanism.

Next, we present the auction-based  mechanism for the allocation of the  demand requested by liquid agents to be produced by buffer agents   that we name  the \textit{Spillover Algorithm}.
In each iteration,  liquid and buffer agents negotiate for the allocation of the liquid demands in the buffers through multiple auctions (one auction per each buffer) in which each buffer agent announces its available resources and liquid agents bid for available buffers  with locally lowest cost. Then, each buffer agent allocates liquid agents' demand  that locally maximizes its  social welfare.

\subsection{Spillover Algorithm: Liquid Agent}
We propose next a decentralised anytime algorithm  for liquid   agents.
Note that an anytime algorithm should be stoppable at any time, providing a feasible solution to  the problem (as an approximation to the optimal solution). Also, the provided solution should monotonically improve with the runtime, contrary to  Lagrangian relaxation  that may oscillate between two points.
This is why we explore a heuristic approach that follows given rules in iteratively allocating liquid agent demands to buffers based on the spillover effect.
The basic idea here is that the liquid demand that cannot be allocated in a   buffer where it appears due to its limited capacity, spills over through its dedicated tube towards other neighbouring buffers. The direction and quantity of spillage will depend on the accumulated  values of the production, setup, holding and back order cost throughout the planning time horizon  and the backlog cost after the end of the time horizon and their relative values in respect to other concerned liquids.
%
%

We introduce the definitions that are the building blocks   of the algorithm.
\begin{definition}
 \textit{Accumulated unit production cost} $UPC_{ak}$ of liquid agent $a = (i,t) \in A$ for each  $k \in T \cup \{|T|+1\}$:
\begin{equation}\label{UPC-itk}
    UPC_{ak} = \begin{cases}
       c_{ik}+s_{ik}+\sum_{m=k}^{t-1}h_{im}, & \forall k\ \in T | k < t\\
       c_{ik}+ s_{ik}, & \text{for } k=t\\
        c_{ik}+s_{ik} + \sum_{m=t+1}^{k}b_{im}, & \forall k \in T | k > t\\
        M\cdot \sum_{m=t+1}^{k} b_{im}, & \text{for } k=|T|+1
        \end{cases}
  \end{equation}
  \end{definition}
That is, $UPC_{ak}$  for period $k$ is composed of a setup ($s_{ik}$) and  production cost ($c_{ik}$) and  the accumulated  holding and back order costs (if any)  depending on   the relation of period $k$ with the demand period $t$ and with the time horizon $T$. Even though $c_a$ (agent production cost) and $s_{ak}$ (agent's set-up cost) enter the objective function  (\ref{minPiL})  very differently (one is multiplied with $u_a^k$, the other with $y_ak$), to achieve a straight-forward computation of $UPC$ and avoid iterative convergence issues, we model it as the overall cost of a unitary item production in period $k$ for agent $a$.

%
Assuming non-negative cost parameters, surplus (unsold) stock $x_{i,|T|+1}^+$ at period $|T|+1$ in problem $P$ (\ref{minPiG})--(\ref{variableGT1P}) will be zero. Since we want to avoid backlog at period $|T|+1$, accumulated unit production costs for period $|T|+1$ in the proposed Spillover Algorithm are found by summing back order costs through a given time horizon   multiplied by    a very high number $M$. $M$ is a design parameter, whose value is given by the owner of a liquid agent $a \in A$. If its value is relatively low, an optimal solution may contain  some unmet demand in a given time horizon.

\begin{definition}
\textit{Estimated accumulated  cost} $EAC_a$ of liquid agent $a=(i,t)$ is accumulated unitary cost  made of unitary production, setup, holding, and  back order   costs estimated over available periods $k \in \Gamma_a  \cup \{|T|+1\}$, where $\Gamma_a \subset T$ is the set of available buffers that have capacity to produce at least one unit of its product:
 \begin{equation}\label{ROC}
   EAC_a=\sum_{k \in \Gamma_a \cup \{|T|+1\}} UPC_{ak}
 \end{equation}
 \end{definition}

The  decision-making procedure for each liquid agent  $a\in A$ is presented in Algorithm \ref{alg:spillover_algorithmApprox}.
\IncMargin{0em}
\begin{algorithm}[h]
compute $UPC_{ak}$ for all $k \in T \cup \{T+1\}$\\
transmit $d_a$, $r_i$ to all $k \in T$ \\
receive $R_k$ from all $k \in T$ \\
\While{$d_a\geq 1$ and  $\big\{\lfloor (\max_{k \in T} R_k)/r_i \rfloor\big\}\geq 1$}{ \label{algL:while}
create set  $\Gamma_a \subset T$ of available buffers such that $k \in \Gamma_a$ iff $\lfloor (\max_{k \in T} R_k)/r_i \rfloor\geq 1$ \label{algL:gamma}\\

sort  $\Gamma_a$ based on non-decreasing $UPC_{ak}$ value \label{algL:UPC}\\
compute estimated accumulated cost $EAC_a$ \label{algL:EAC}\\
  \ForEach{$k \in \Gamma_a$} { \label{algL:foreach}
$\rho_{ak}=
                                \begin{cases}
                                 \min (\lfloor R_k/r_i\rfloor \cdot r_i, d_a \cdot r_i)  \text{ if } k=t,\\
                                \min\big(\lfloor R_k/r_i \rfloor  \cdot r_i, d_a \cdot r_i-
                                \sum_{m=\Gamma_{a1}}^{\Gamma_{a, k-1}} \rho_{am}\big), otherwise;
                                \end{cases}$ \label{algL:dak}\\
        transmit bid $B_{ak}=(\rho_{ak}, EAC_a)$ to  $k$ \label{algL:bid}
        }
receive $u_a^k$ for all $k \in \Gamma_a$ \label{algL:rec_u}\\
$d_a= d_a- \sum_{k \in \Gamma_a}u_a^k$ \label{algL:da}\\
 transmit $d_a$ to all $k \in T$ \label{algL:trd}\\
 receive $R_k$ from all $k \in T$  \label{algL:rec_Rk}\\
}
$\xmeno_{a,|T|+1}=d_a$;\\
compute  $Z_a(\mathbf{u}_a)$; \\
return    $Z_a(\mathbf{u}_a)$ and $\xmeno_{a,|T|+1}$
\caption{Spillover Algorithm:   liquid agents $a=(i,t) \in A$ }
\label{alg:spillover_algorithmApprox}
\end{algorithm}\DecMargin{0em}
The algorithm initiates with  transmitting  unaccommodated demand $d_a$ and resource requirement per product ($r_i$) to buffer agents $k \in T$, and receiving buffers' capacities $R_k$.
If none of the buffers has available capacity (line \ref{algL:while}), the procedure terminates after computing and returning its unmet demand $\xmeno_{a,|T|+1}$ and heuristic approximate  cost $Z_a(\mathbf{u}_a)$, where $\mathbf{u_{a}}=\{u_a^k | k \in T \}$,  given by:
\begin{equation}\label{zObjFuncLiqApprox}
Z_a(\mathbf{u_a}) =  \sum_{k \in T }UPC_{ak}\cdot u_a^k
+UPC_{a,|T|+1}  \cdot \xmeno_{a,|T|+1}
\end{equation}

By summing heuristic costs $Z_a(\mathbf{u_a})$ in  (\ref{zObjFuncLiqApprox}) over all liquid agents $a \in A$, we can easily obtain the overall system approximate heuristic cost made of the heuristic costs of the allocation of the demands of all items over all periods.
Otherwise, liquid agent with positive demand ($d_a\geq 1$) creates a set of available buffers $\Gamma_a$ (line \ref{algL:gamma}).
Then, it computes $UPC_{ak}$ for each buffer $k \in \Gamma_a$ and orders the buffers with available capacity in   $\Gamma_a \subset T$  based on the non-decreasing $UPC_{ak}$ value (line \ref{algL:UPC}).
$EAC_a$ is computed on line \ref{algL:EAC} and is used as a part of  bid $B_{ak}$ sent to each  of the buffer agents $k \in \Gamma_a$ considering   resource demand $\rho_{ak}$ and  available capacity $\lfloor R_k/r_i \rfloor$  until the extinction of the unaccommodated demand $d_a$ or the available capacity (line  \ref{algL:dak}), while
$\Gamma_{a1}$  and $\Gamma_{a, k-1}$ are the $1^{st}$ and $(k-1)^{st}$ element of $\Gamma_a$.
Liquid agent sends this information as a part of bid $B_{ak}=(\rho_{ak}, EAC_a)$   to each buffer $k \in \Gamma_a$ (line \ref{algL:bid}).
%
The direction and quantity of spillage will depend on $EAC_a$. 

By communicating its bid in terms of resource demand $\rho_{ak}$ for each one of the elements $k \in \Gamma_a$ and the value of $EAC_a$, a liquid agent does not have to disclose its sensitive private information regarding the values of  unitary production, holding, setup   and back order cost in each time period of the planning time horizon as well as the unitary cost of backlog after the end of the planning time horizon   nor reasons for its decision-making.
We choose this greedy strategy of producing as much as possible starting with the first buffer agent in set $\Gamma_a$   since we aim to minimize agent's estimated accumulated cost $EAC_a$ for the rest of available periods in the horizon.
Then, on receiving  production response $u_a^k$ (line \ref{algL:rec_u}) from buffers $k \in \Gamma_a$, liquid agent $a$ updates its unaccommodated demand (line \ref{algL:da}) and transmits it to all buffers  $k \in \Gamma_a$ (line \ref{algL:trd}).
After receiving available production capacities $R_k$ from all $k \in T$ (line \ref{algL:rec_Rk}), if there still remains an unaccommodated demand   $d_a$ and available capacities $ \lfloor R_k/r_i \rfloor$,   new iteration starts; otherwise, the algorithm terminates.


\subsection{Spillover Algorithm: Buffer Agent}
Each buffer agent $k \in T$ knows its capacity and acts as an auctioneer for accommodation of liquid agent bidders' demand for period $k$. It orders all liquid agents bids in a non-increasing order and greedily  accommodates for production   $\mathbf{u_{k}}=\{u_a^k | a \in A\}$ allocating as much production resources as possible to the liquid agent bidder with the highest $EAC_a$ value up to the depletion of its available production capacity $R_k$.


\begin{algorithm}

transmit  $R_k$ to all $a \in A$; \label{algC:TrRk}\\
receive $d_a$, $r_i$  for all $a \in A$  \label{algC:Rxdr}\\
$A^u= \{a \in A | d_a >0\}$   \label{algC:Au}\\
\While{$|A^u|\geq 1$ and $\big\{\lfloor R_k/\min_{a \in A^u}r_i \rfloor \big\}\geq 1$}{   \label{algC:while}
receive $B_{ak}= (\rho_{ak}, EAC_a)$ for all $a \in A^u$   \label{algC:RxB}\\
create ordered set $BA_k=\{(a,\rho_{ak})| a \in A^u\}$  based on non-increasing $EAC_a$ value   \label{algC:BA}\\
// allocate liquid agents in $BA_k$ for production\\
 \ForEach{$a \in BA_k$}{   \label{algC:foreach}
 $u_a^k = \min ( \rho_{ak}/r_i, \lfloor R_k/r_i \rfloor)$   \label{algC:u_ak}\\
 $R_k = R_k-u_a^k \cdot r_i$  \label{algC:updateRk} \\
    }
     transmit $u_a^k$ and $R_k$ to all $a\in A^u$   \label{algC:TrU}\\

receive $d_a$ from all $a \in A^u$   \label{algC:Rxda}\\
  $A^u= \{a \in A | d_a >0\}$ \label{algC:updateAu}\\
}
return   $u_a^k$,  $\forall a \in A$
\caption{Spillover Algorithm: buffer agents $k \in T$} 
\label{alg:spillover_algorithm-containerApprox}
\end{algorithm}

The decision-making   of each buffer agent runs in iterations (Algorithm \ref{alg:spillover_algorithm-containerApprox}).
Initially, it transmits  its available capacity $R_k$ (line \ref{algC:TrRk}) and receives demands $d_a$ and resource requirements per product $r_i$ of all liquid agents $a=(i,t) \in A$ (line \ref{algC:Rxdr}).
Then, it obtains the set $A^u$ of liquid agents with unmet demand (line \ref{algC:Au}).
If there is unmet demand and sufficient production capacity (line \ref{algC:while}), after receiving bids from bidding liquid agents (line \ref{algC:RxB}), it creates an ordered tuple of bidding agents $BA_k$ based on non-increasing $EAC_a$ value (line \ref{algC:BA}).
This policy minimizes the maximum cost of the liquid agents bidding for a buffer and can be seen as a fairness measure to increase egalitarian social welfare among liquid agents.
Then, it    allocates the highest possible production level to liquid agents in $BA_k$  considering resource demand $\rho_{ak}$ and remaining  capacity   (line \ref{algC:u_ak}) and transmits the production values  and remaining capacity to concerned agents $a \in A^u$  (line \ref{algC:TrU}).
%
On receiving unaccommodated demand $d_a$ from   $a \in A^u$ (line \ref{algC:Rxda}), the buffer updates $A^u$ and checks whether the termination conditions on unaccommodated demand or its remaining capacity are fulfilled (line \ref{algC:while}). If so, it terminates; otherwise,  it repeats.

\subsection{Spillover Algorithm analysis}
This section  analyses the algorithm's termination, optimality, complexity and the protection of sensitive private information.

\paragraph{Termination.}
The Spillover Algorithm terminates in at most $|I|\cdot|T|^2$ iterations since it iterates through $|I|\cdot|T|$ liquid agents and each of them does at most $|T|$ iterations. It stops for a liquid agent $a \in A$ in the worst case when all the periods of the time horizon $T$ have been bid for.
This is guaranteed to occur in  $|T|$ steps,  if the time horizon is bounded. If agent set is bounded, in the worst case when applied serially, this will happen in $|A|\cdot|T|$ steps for all agents.
If the  overall demand through the given time horizon
may be eventually allocated for production, i.e., if
$\sum_{a \in A} d_a \leq \sum_{k \in T}\lfloor R_k/r_i \rfloor$,
no unmet demand (big M)  cost will be accrued.
\paragraph{Optimality.} Whether the complete allocation obtained upon termination of the auction process is optimal depends strongly on the method for choosing the bidding value. The heuristic of ordering local
$EAC_a$ values does not have a guarantee of a global optimum solution but guarantees local optimum. Thus, in case of an unpredicted setback or change in capacities or demands, instead of updating the plan for the whole system, as is the case in the centralised control, it is sufficient to modify only the plans of the directly involved agents. 

\paragraph{Complexity  of  Algorithm \ref{alg:spillover_algorithmApprox} (Liquid agent).}
A liquid agent does at most $|T|$ iterations in the worst case (line \ref{algL:while}) with the aim to accommodate its demand  in any of the buffer agents.
A bid from liquid agent $a$ is either (i) completely satisfied, (ii) partially satisfied or (iii) completely unsatisfied. In case (i), it may be possible that a liquid agent with still remaining unallocated demand bids again for the same buffer (if its demand allocation request was  refused in other buffers). The buffer may i) allocate all the liquid agent's demand up to the buffer's capacity or up to the extinction of the liquid agent's demand or ii) reject the allocation since it is already full.
Thus, although liquid agent $a$ may bid at most two times for each buffer, in each iteration, at least one buffer is filled up and thus discarded  for the next round of bids of the same liquid agent.
The calculation of $EAC_a$ (line \ref{algL:EAC}) requires $|T|$ iterations in the worst case as well as sending bids to buffer agents (foreach loop in line \ref{algL:foreach}).
Thus, in each   loop in line \ref{algL:while}, liquid agents send $2 \cdot |T|$ messages to and receive $2 \cdot |T|$ messages  from buffer agents (lines \ref{algL:rec_u} and \ref{algL:rec_Rk}); exchanged messages $O(|T^2|)$.

\paragraph{Complexity of   Algorithm \ref{alg:spillover_algorithm-containerApprox} (Buffer agent).}
The main loop (line \ref{algC:while}) of a buffer agent is repeated $2 \cdot |A|$ times in the worst case, since in every iteration a new bid (different from a previous one) from a liquid agent could be received.

In each iteration, a buffer agent sends (lines \ref{algC:TrU}) and receives  $2 \cdot |A|$ messages (lines \ref{algC:RxB} and \ref{algC:Rxda}) in the worst case, respectively. Thus, the complexity in number of messages exchanged is $O(|A|)$. Note that sorting the received bids $BA_k$ (line \ref{algC:BA}) can be done in $O(|A| \cdot log |A|)$. However, this is done locally, not being as complex as exchanging $O(|A|)$ messages.
In total, the complexity of the decision-making algorithm of each buffer agent is $O(|A|^2)$.

\paragraph{Protection of sensitive private information.}
In the Spillover Algorithm, the most  sensitive private  information includes the values of  unitary production $c_{it}$, holding $h_{it}$, setup $s_{it}$   and back order cost $b_{it}$ in each time period $t$ of the planning time horizon $T$ as well as the unitary cost of backlog $b_{i,|T|+1}$ after the end of the planning time horizon.
None of this information is shared among liquid agents representing users of production resources.
To be able to learn deterministically the cost values of a liquid agent by a buffer agent from the Estimated Accumulated Cost $EAC_a$, at least ($4|T|+1$) linearly independent $EAC_a$ values must be available (a solution to $n$ unknowns can be found deterministically by solving $n$ linearly independent equations). Here, $4|T|$ refers to the number of cost parameters ($c_{it}, h_{it}, b_{it}, s_{it}$) over  $|T|$ time periods and we need 1 additional equation for finding the value of $M$.
Note that there are at most $|T|$ possible bids  in each iteration and at most 2 bids per buffer made by a liquid agent.
Since deterministic inference of the cost parameter values and a very large number $M$ in the Spillover Algorithm is possible only if $4|T|+1$ linearly independent $EAC_a$ values are available, it is impossible to obtain the sensitive private cost parameter values of liquid agents by buffer agents in the Spillover Algorithm.

\section{Simulation experiments}\label{experiment}
In this section, we  compare the performance of the proposed Spillover Algorithm and the centralised  and optimal CPLEX solution  considering randomly generated and diversified  problem set for large-scale capacitated lot-sizing. To the best of our knowledge, there are no other decentralised solutions to the MCLSP-BOSI problem to compare with. We tested a general DCOP ADOPT-n (\cite{ModiSTY05}) but due to its modelling for n-ary constraints and the presence of mostly global constraints in MCLSP-BOSI, with only 4 periods and 4 liquid agents, it didn't find a solution in a reasonable time.

\subsection{Experiment setup}
The experiment setup was performed   following the principles in \cite{diaby1992lagrangean} and \cite{giordani2013distributed}. %
We consider the case of \textit{Wagner-Within costs}, i.e., the costs that prohibit speculative motives for early and late production resulting in speculative inventory holding and back-ordering, respectively. Considering constant production costs $c_a$ for all $a=(i,t) \in A$, setup, holding, and back order costs comply with the assumption of Wagner-Within costs, i.e.,
\begin{equation}\label{Wagner-WithinSetupCosts}
-b_{ik}<s_{i,k+1}-s_{ik}<h_{ik}, \; \forall k \in T
\end{equation}
The value of  $s_{i1}$ for each item $i \in I$ is chosen randomly from the uniform distribution in the range [50, 100], i.e. $s_{i1}\sim U(50,100)$, and the value of the setup cost of every following period $\{s_{i2}, \ldots, s_{i,|T|}\}$ is computed by the following formula $s_{i,k+1}=s_{ik} + X$, where 
$X \sim U(-1,1)$.

For simplicity, holding costs $h_{ik}$ are assumed constant in time, i.e., $h_{ik}=h_i$, and are generated randomly from $U(20,100)$ for each item $i \in I$. Backorder costs are computed from the holding costs considering a multiplication factor of 10, 2, 0.5, and 0.1 (the obtained values are rounded to have all integer values).
When considering backlog costs in (\ref{UPC-itk}),  to produce as much as possible during the time horizon, we model $M$ as a very large number whose value is 10000.
We assume   unitary production cost $c_{it}$  for each item $i \in I$  to be constant in time, i.e. $c_{it}=c_i$, and $c_i \sim U(1000, 10000)$.

%
%
In the experiments, we consider a large  production system scenario producing from 50 to 150 items with increase 10, over a time horizon composed of 100 periods with the value of mean item demand per period  
$\bar{d}_i\sim U(100,1000)$. This sums up to overall demand from 2.75 to 8.25 million units on average in the given time horizon.
%

For each item $i \in I$, demand $d_{it}$ is generated from the normal distribution with mean $\bar{d}_i$ and standard deviation $\sigma = \bar{d}_i/\kappa$, with $\kappa>1$, i.e. $(d_{it} \sim N[\bar{d}_i, $ $(\bar{d}_i/\kappa)^2])$. 
Standard deviation of the distribution of the item-period demands controls the demand variability. We experiment two levels of demand variability:   high level whose  $\sigma = \bar{d}_{it}/2$ ($\kappa=2$) and   low level with $\sigma=\bar{d}_{it}/4$ ($\kappa=4$). The larger the value of $\kappa$, the more ``lumpy'' the demand, i.e., the greater the differences between each period's demand and the greater the number of periods with zero or fixed  base level demand  (\eg   \cite{wemmerlov1984lot}).
%
%
%
%
%
%
For simplicity, but w.l.o.g, the values of $x_{i0}^{+}$, $\xmeno_{i0}$, and $y_{i0}$  are assumed to be equal to zero.
Resource requirement $r_i$ for production of a unit of $i \in I$ is generated 
from $U(1,3)$.
%
We consider variable production capacities generated from 
 $(R_k \sim N[110000, (110000/4)^2])$.
This value is chosen to satisfy the overall demand of $|I|=100$ items with average demand  $\bar{d}_i=550$ units with average resource requirements $\bar{r}_i=2$ for all $i \in I$ over 100 periods.
%

As the   key performance indicators, we consider average computational time and optimality gap  compared  between the solution of the Spillover Algorithm implemented in Matlab and the optimal solution computed with IBM ILOG CPLEX Studio 12.8 with CPLEX  solver, both run on an Intel Core  i5 with 2.4 GHz and 16 GB RAM.
%
We tested  scenarios, from very decongested ones with   the overall demand    representing $50\%$ of the available production resources in the horizon, to very congested scenarios where this percentage increases to $150\%$. We combine the 11 cases of demand  varying from 50 to 150 items with increment of 10 over 100 periods with 2 cases of   variability and 4 cases of back order costs related to the holding costs, resulting in a total of 88 experiment setups. 

\subsection{Results}
\begin{figure}
\begin{center}
\includegraphics[width=0.8\columnwidth]{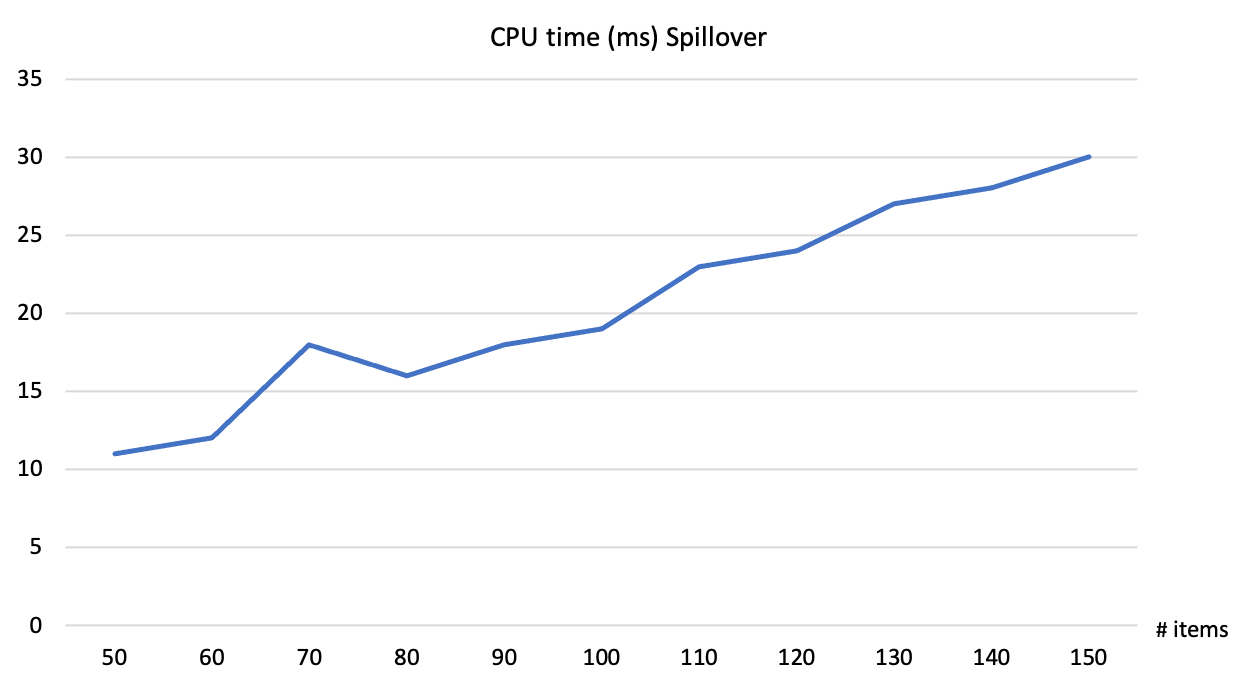}
\caption{Average computational time of the Spillover Algorithm for different numbers of items}
\label{CPUTimeSpillover}
\end{center}
\end{figure}

\begin{figure}
\begin{center}
\includegraphics[width=0.9\columnwidth]{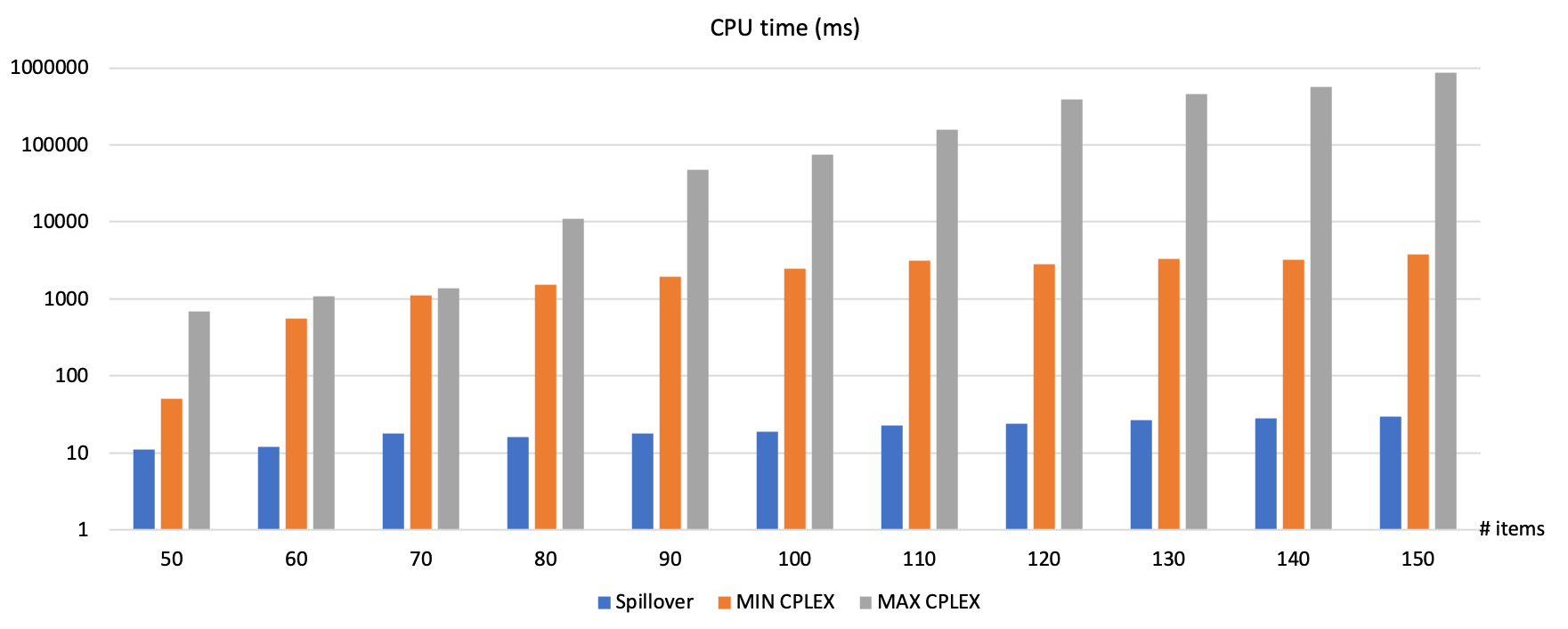}
\caption{Comparison of the average computational time (in logarithmic scale) of the Spillover Algorithm with the minimum and maximum computational times of CPLEX solver}
\label{CPUTime_SpilloverCPLEX}
\end{center}
\end{figure}

 The simulation results of the Spillover Algorithm
with the approximation scheme are shown in Table \ref{Table:SpilloverApproximationResults}.
The table contains the optimality gap (average of 8 experiments each, as described previously) in relation to the solution found in CPLEX 
together with their computational times.
The optimality gap is the one used by CPLEX and is obtained as $(z_x - z_o)/z_o$, where $z_x$ is the cost of the solution found by the Spillover Algorithm and $z_o$ is the cost of the CPLEX optimal solution.
The mean gap value is $25\%$ and the individual gap value throughout the instances is rather constant and independent of the item number. This gap value, for such a heuristic approximation approach, is a very good result considering that heuristic approaches generally do not have quality of solution guarantees.
The computational time of CPLEX has high variability for the same problem size.  Thus, Table \ref{Table:SpilloverApproximationResults} shows the minimum and maximum execution times (in seconds) for each number of items.
The Spillover Algorithm execution time is less than 0.1 sec for all the experiments and grows lineally with the number of items (see Figure \ref{CPUTimeSpillover}).
This confirms its good performance in relation to CPLEX, whose computational time increases exponentially with the increase of the item set size and thus scales poorly.
Figure \ref{CPUTime_SpilloverCPLEX} shows graphically, using a logarithmic scale, the minimum and maximum times taken by CPLEX for different problem sizes (number of items) and the average Spillover Algorithm time.
However, the emphasis  here is not on proposing an optimal method in a centralised  environment, but to decentralise the coordination decisions of production planning in environments where the exposure of private and sensitive information is desirably minimized while still having a reasonably good global solution away from the control of a centralised decision maker.


\begin{table}
    \centering
    \caption{Summary of the computational results}
    \label{Table:SpilloverApproximationResults}
\begin{tabular}{c|c|c|c|c|c|c|c|c|c|c|c} \toprule
    \# items  & 50 & 60 & 70 & 80 & 90 & 100 & 110 & 120 & 130 & 140 & 150 \\
    \midrule
    Avg. Gap (\%)  & 28 & 28 & 30 & 16 & 16 & 28 & 27 & 28 & 27 & 27 & 27 \\
    Spillover (ms)& 11 & 12 & 18 & 16 & 18 & 19 & 23 & 24 & 27 & 28 & 30 \\
    Min CPLEX (s)& 0.05 & 0.56 & 1.1 & 1.53 & 1.92 & 2.49 & 3.1 & 2.83 & 3.35 & 3.2 & 3.75 \\
    Max CPLEX (s)& 0.69 & 1.07 & 1.39 & 11 & 47 & 74 & 157 & 391 & 451 & 571 & 869 \\
  \bottomrule
\end{tabular}
\end{table}

\section{Discussion and conclusions}\label{Conclusions}
Open and shared factories are becoming popular in Industry 4.0 as another component of today's global economy. In such facilities, production resources and product owners coexist in a shared environment.
One of the main issues faced by product (item) owners  that compete for limited production resources held by multiple resource owners in such factories is the exposure of their private and sensitive information including the values of  unitary production, holding, setup   and back order cost in each time period of the planning time horizon as well as the unitary cost of backlog after the end of the planning time horizon. 

The scope of this paper was not to propose a more computationally efficient heuristic for centralised MCLSP-BOSI problem, but to study a decentralised version of the MCLSP-BOSI problem and develop a heuristic approach  applicable to  intrinsically decentralised shared and open factories. Centralised state-of-the-art heuristics cannot be applied to this problem with  self-concerned and individually rational  resource and item owners since these heuristics require complete exposure of everyone's  private and sensitive information.

Therefore, in this paper, we presented  a decentralised and dynamic  variant of the classic MCLSP-BOSI problem with time-dependent costs. To reach a decentralised variant of the problem and to control locally the linear increase of product owners' costs, we decomposed the problem  based on \textit{item-period} pairs.

As a solution approach to the decentralised and dynamic  MCLSP-BOSI problem, we proposed the \textit{Spillover Algorithm}, to the best of our knowledge,  the first heuristic decentralised algorithm for this problem that complies with the intrinsically decentralised nature of  large and shared open factories.
A heuristic solution was needed to cope with the NP-hard nature of the (decentralised) MCLSP-BOSI problem.
The spillover heuristic is formulated as a multi-agent algorithm in a liquid flow network model with buffers:    each item-period pair  is represented by a \textit{liquid agent} responsible for obtaining production resources (robots) to manufacture its demand, \ie product of type $i$ requested  to be produced  by the means of  bidding for resource allocation at time $t$.
Likewise, production capacity  (number of available robots) in each period  is represented by a \textit{buffer agent} responsible for allocating the capacity  to bidding \textit{liquid agents}.

An auction-based algorithm leveraging spillover effect has been designed for  a one-on-one negotiation between the liquid agents requesting item production and their buffer agents of  interest.
Each liquid agent sends greedy bids (consisting of the amount of the item to be produced, and the agent's estimated accumulated cost for available buffers) to the buffer agents in order of non-decreasing accumulated unit production cost, and each buffer agent greedily accepts bids in order of the bidding liquid agents' non-increasing estimated accumulated cost for all buffers. This is repeated until all demands are allocated for production throughout the planning time horizon.

The proposed multi-agent auction-based approach  has the advantage that agents do not need to reveal all their private and sensitive information (\ie unitary costs of production, setup, back order, and holding) and that the approach is decentralised.
Distributed problem solving here includes sharing of estimated accumulated costs for each item-period pair, \ie  a heuristic  unitary  demand cost accumulated over time periods that are still available for resource allocation. The latter can be interpreted as an approximate  ``resource conflict price'' paid if
a unit of demand is not allocated for production in the requested period.

Note that the state-of-the-art heuristics are centralised solution approaches that are not applicable in intrinsically decentralised open and shared factories with self-interested and individually rational resource and product owners. Thus, the comparison of the performance of the Spillover Algorithm with these solution approaches is meaningless.

Since all agents in the Spillover Algorithm have only a local view of the system, a disadvantage is that global optima are not necessarily achieved with such a decentralised control.

We presented an experimental evaluation with randomly generated instances that nonetheless shows that the solutions obtained using the spillover effect heuristic are only about 25$\%$ more expensive than the optimal solutions calculated with CPLEX.
The average optimality gap values are between 16$\%$ and 30$\%$ for  the 11 tested  item type numbers (from 50 to 150 different items (products)), where the average has been taken over 8 randomly generated instances for each choice of the number of item types.
However, the Spillover Algorithm gives an anytime feasible solution running in close-to real time, under 30 milliseconds, while CPLEX requires between 4 and 869 seconds for the largest tested instances.

The low computational complexity of the Spillover Algorithm facilitates the implementation in shared and open factories with competitive stakeholders who want to minimize their sensitive and private information exposure.

The Spillover algorithm   can be applicable, with no substantial changes, to other domains where multiple independent self-concerned agents compete for limited shared resources distributed over a  time horizon of a limited length. This is the case, for example, in flight arrival/departure scheduling to maximize runway utilization and the allocation of electric vehicles to a charging station. In the former, flight scheduling case, liquid agents (representing flights) compete for runway time slots, which are managed by buffer (runway) agents responsible for allocating their landing/take-off capacity in each time period. At peak hours, sometimes it is inevitable that some flights change their schedules because of the limited capacities   of the runways. Thus, dynamic (re-)allocation is required. In the application of the Spillover algorithm to this case, the production cost is the time slot cost for an aircraft, back order costs reflect the consequences of flight delays (e.g. compensations to clients, damage of reputation, etc.), and holding costs represent the monetary effect of early departures/arrivals. Setup costs are the operational costs and other charges for an aircraft using the runway at certain time period.

The Spillover Algorithm is resilient to crashes as the buffer agents continue to assign their resources as long as there is at least one liquid agent bidding for them. This topic will be treated in future work.
In future work, we will also study  more accurate spillover rules that iteratively approximate estimated accumulated costs as the bidding progresses while guaranteeing  convergence and the protection of sensitive private information.
Furthermore, in the Spillover algorithm, buffer agents prioritize bids from liquid agents  based on their estimated accumulated cost $EAC$ value included in each bid. The higher the value, the higher the priority for resource allocation of a liquid agent. Thus, a strategic  liquid agent may want to report a falsely high $EAC$ value to get the production resources at requested time periods.
Open issues of strategic agents, trust and incentives not to lie about production demand, estimated cost parameters, and production capacities will also be studied in future work.
Examples of lines to explore are game theory and mechanism design (e.g. Vickrey-Clarke-Groves mechanism).

\section*{Acknowledgements}
This work has been partially supported by the ``E-Logistics'' project
funded by the French Agency for Environment and Energy Management (ADEME) and by the ``AGRIFLEETS'' project ANR-20-CE10-0001 funded by the French National Research Agency (ANR) and by the STSM Grant funded by the European ICT COST Action IC1406, ``cHiPSeT'',
and by the Spanish MINECO projects RTI2018-095390-B-C33 (MCIU/AEI/FEDER, UE) and TIN2017-88476-C2-1-R.

\bibliographystyle{model4-names}

\bibliography{RCIM-database}
\end{document}